\documentclass[conference,a4paper]{IEEEtran}
\IEEEoverridecommandlockouts

\usepackage[hidelinks]{hyperref}
\usepackage[cmex10]{amsmath}
\usepackage{amssymb,amsfonts}
\usepackage{dblfloatfix}
\usepackage{fixltx2e} 

\usepackage[pdftex]{graphicx}
\graphicspath{{Figures/}}

\usepackage{booktabs}
\usepackage{siunitx}
\usepackage[numbers,compress]{natbib}
\usepackage{bm,bbm}
\usepackage{orcidlink}
\usepackage{xcolor}
\usepackage{multirow}
\usepackage{algorithm}
\usepackage{algpseudocode}

\usepackage{fancyhdr}

\fancypagestyle{IEEEfooter}{
    \fancyhf{}
    \fancyfoot[L]{%
        \scriptsize 979-8-3195-0862-1/26/\$31.00 \copyright 2026 IEEE%
    }
    \fancyfoot[C]{\scriptsize\thepage}
    \fancyfoot[R]{\scriptsize IGARSS 2026}

}

\begin{document}

\title{FiLM-GPNet: Geometry-Aware Pseudo-Supervised Phase Restoration with Zero-Shot Generalization for Large Temporal InSAR Stacks}

\author{
\IEEEauthorblockN{Getnet Demil$^{*}$\orcidlink{0009-0006-5158-0895}}
\IEEEauthorblockA{\textit{University of Oulu}\\
Oulu, Finland\\
getnet.demil@oulu.fi}
\and
\IEEEauthorblockN{Muhammad Farhan Humayun$^{*}$\orcidlink{0000-0002-1160-813X}}
\IEEEauthorblockA{\textit{University of Turku}\\
Turku, Finland\\
farhan.humayun@utu.fi}
\and
\IEEEauthorblockN{Tomi Westerlund}
\IEEEauthorblockA{\textit{University of Turku}\\
Turku, Finland\\
tovewe@utu.fi}
\and
\IEEEauthorblockN{Jukka Heikkonen}
\IEEEauthorblockA{\textit{University of Turku}\\
Turku, Finland\\
jukhei@utu.fi}
\and
\IEEEauthorblockN{Mourad Oussalah}
\IEEEauthorblockA{\textit{University of Oulu}\\
Oulu, Finland\\
mourad.oussalah@oulu.fi}
\thanks{$^{*}$These authors contributed equally to this work.}
}

\maketitle
\thispagestyle{IEEEfooter}
\begingroup
\renewcommand\thefootnote{\fnsymbol{footnote}}
\footnotetext{%
\emph{Code $\&$ Supplementary material available at:}
\href{https://github.com/getnetdemil/FiLM-GPNet-InSAR-Phase-Denoising}{\textbf{GitHub Repo}},
\href{https://github.com/getnetdemil/FiLM-GPNet-InSAR-Phase-Denoising/blob/main/Supplemntary_material.pdf}{\textbf{Appendix}}.}
\endgroup

\begin{abstract}

The growing availability of dense commercial Synthetic Aperture Radar (SAR) time series enables temporal Interferometric SAR (InSAR) analysis, but fixed classical filters fail under heterogeneous acquisition geometries, degrading phase quality and temporal consistency. We propose \textbf{FiLM-GPNet}, a geometry-conditioned network for wrapped-phase restoration that explicitly adapts to acquisition differences using Feature-wise Linear Modulation (FiLM) and a 7D per-pair geometry descriptor. The model is trained with pseudo-supervision from Goldstein-filtered interferograms and regularized by interferometric physics via triplet-closure consistency, while also estimating per-pixel aleatoric uncertainty. Experiments on three Capella Spotlight stacks from the IEEE GRSS 2026 Data Fusion Contest show that FiLM-GPNet reduces temporal residual by \textbf{68\%} (Hawaii) and \textbf{66\%} (Western Australia) relative to the Goldstein baseline, alongside closure error reductions of \textbf{10\%} and \textbf{13\%}, respectively. In Western Australia, it further improves unwrapping success rate by \textbf{7.7} percentage points and Digital Elevation Model (DEM) Normalized Median Absolute Deviation (NMAD) by \textbf{31\%}. The model also shows strong zero-shot generalization to a geographically and geometrically distinct third stack (Los Angeles) without retraining, supporting geometry-conditioned restoration as an effective alternative to fixed classical filtering across heterogeneous stacks.\\
\end{abstract}


\section{Introduction}

Dense temporal stacks from commercial X-band SAR constellations expose a fundamental limitation of classical InSAR processing: geometry-blind filters such as Goldstein apply essentially uniform smoothing across interferograms despite substantial variation in temporal baseline, perpendicular baseline, incidence angle, and viewing geometry \cite{goldstein1998radar, baran2003modification, 11186241}. In heterogeneous stacks, this often leads to inconsistent interferograms, elevated triplet-closure errors, unstable unwrapping, and weaker multi-temporal inversion performance. At the same time, learning-based phase restoration remains difficult to supervise in operational settings because clean reference interferograms are generally unavailable, and many existing approaches rely on synthetic, proxy, or semi-supervised supervision \cite{9957081, 11000282}.

To address this problem, we propose \textbf{FiLM-GPNet}, a geometry-conditioned pseudo-supervised network for phase restoration in large temporal InSAR stacks. Our main contributions are:
\begin{itemize}
    \item A FiLM-conditioned U-Net backbone that adapts phase restoration to per-pair acquisition geometry instead of applying a fixed filter across heterogeneous stacks.
    \item A pseudo-supervised training strategy that uses Goldstein-filtered interferograms as proxy targets and regularizes learning through triplet-closure consistency.
    \item Per-pixel aleatoric uncertainty prediction for downstream confidence assessment.
    \item Zero-shot generalization: a model trained on one geographic stack transfers without retraining to a geographically and geometrically distinct site.
\end{itemize}

We evaluate the proposed approach on multiple stacks from the IEEE GRSS 2026 Data Fusion Contest dataset.

\section{Method}
Figure A1 in the appendix summarizes the overall processing pipeline. The main components are described below.

\subsection{Geometry-Aware Pair-Graph Construction}

\begin{figure*}[t!]
  \centering
    \includegraphics[width=0.85\textwidth]{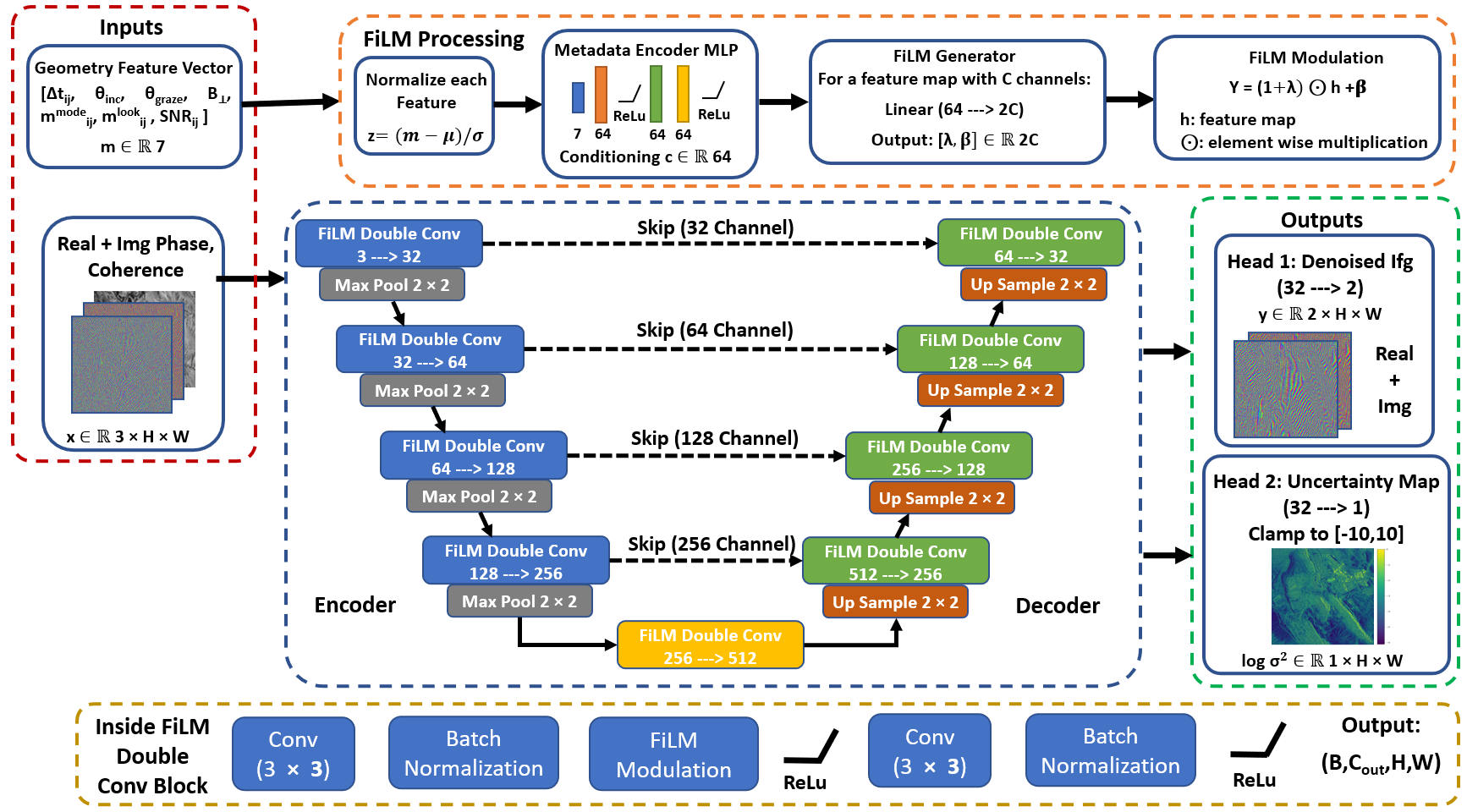}
    \caption{The FiLM-GPNet architecture: input/output blocks, FiLM-based encoder/decoder network and MLP for processing of geometry feature vector.}
    \label{FiLMUNet-Full}
\end{figure*}

\begin{equation}
Q_{ij}
=
\frac{1}{1+\Delta t_{ij}}
\cdot
\frac{1}{1+|\Delta \theta_{\mathrm{inc},ij}|}
\cdot
G_{ij}
\label{eq:qscore}
\end{equation}
where $G_{ij}$ is an additional geometry-consistency factor that penalizes look-angle mismatch, squint mismatch, range/azimuth sampling mismatch, and low footprint overlap whenever those metadata are available. Closure triplets are then enumerated from the accepted graph for later physics-based supervision.

\subsection{PFA-Aware Coregistration and Classical Baseline}

Because the Capella Spotlight PFA SLC pairs used in this work can exhibit geometry-dependent misregistration and variable interferometric quality, we apply a compact engineering pipeline for precise pairwise alignment before learning-based phase restoration. For each accepted pair, a metadata-derived scene-reference seed is refined on multilooked log-amplitude thumbnails by phase correlation, followed by grid-based local matching with normalized cross-correlation and subpixel refinement \cite{Lewis1995,GuizarSicairos2008}. A quadratic first-pass offset model and a linear residual second pass are then estimated with robust outlier rejection, and the final slave image is obtained by single-step resampling of the composed offsets. Among the candidate solutions, we retain the coregistration with the strongest coherence statistics.

Given master image $M$ and co-registered slave image $S_{\mathrm{coreg}}$, the raw interferogram is formed as:
\begin{equation}
I(r,c)=M(r,c)\,\overline{S_{\mathrm{coreg}}(r,c)}
\end{equation}
The local interferometric coherence is then estimated using the standard normalized complex correlation \cite{Zebker1992,Hanssen2001}:
\begin{equation}
\gamma(r,c)=
\frac{
\left|\left\langle M\,\overline{S_{\mathrm{coreg}}}\right\rangle_{W}\right|
}{
\sqrt{
\left\langle |M|^2 \right\rangle_{W}
\left\langle |S_{\mathrm{coreg}}|^2 \right\rangle_{W}
}
}
\label{eq:coherence}
\end{equation}
where $\langle \cdot \rangle_W$ denotes local averaging over a boxcar window. In our experiments, pairs with final mean coherence $\bar{\gamma}<0.3$ are rejected from downstream processing. For classical comparison, we retain both the raw wrapped interferogram and the Goldstein-filtered version \cite{goldstein1998radar}:
\begin{equation}
\hat{I}_b=\mathcal{F}^{-1}\!\left(F_b\,H_b\right),
\qquad
H_b=\frac{|F_b|^{\alpha}}{\max\limits_{\mathbf{k}}\left(|F_b(\mathbf{k})|^{\alpha}\right)+\varepsilon}
\label{eq:goldstein}
\end{equation}
where $F_b=\mathcal{F}\{I_b\}$ is the Fourier transform of block $I_b$, $\hat{I}_b$ is the filtered block, and $H_b$ is a normalized spectral weighting mask proportional to $|F_b|^{\alpha}$. Here, $\alpha=0.5$ controls the filtering strength, $\varepsilon$ ensures numerical stability, and the filtered blocks are recombined by overlap-add weighting.

\subsection{Geometry-Conditioned FiLM-GPNet}

Figure~\ref{FiLMUNet-Full} shows the overall architecture of the FiLM-GPNet and its main sub-modules. For each interferometric pair, the network takes as input a three-channel tensor $x_{ij}\in\mathbb{R}^{3\times H\times W}$, whose channels correspond to the real and imaginary parts of the raw interferogram and the coherence map. Each pair is additionally associated with a normalized seven-dimensional geometry vector:
\begin{equation}
z_{ij} = [\Delta t_{ij}, \theta_{\mathrm{inc},ij}, \theta_{\mathrm{graze},ij}, B_{\perp,ij}, m_{ij}^{\mathrm{mode}}, m_{ij}^{\mathrm{look}}, \mathrm{SNR}_{ij}]
\end{equation}
where $\Delta t_{ij}$ is the temporal baseline, $\theta_{\mathrm{inc},ij}$ the incidence angle, $\theta_{\mathrm{graze},ij}$ the grazing angle, and $B_{\perp,ij}$ the perpendicular baseline, while the remaining terms encode acquisition mode, look direction, and an SNR proxy respectively.

The geometry vector is passed through a two-layer MLP that maps the normalized 7-D input to a 64-D conditioning code. This code is injected into every encoder and decoder block through FiLM, which applies channel-wise affine modulation to intermediate feature maps:
\begin{equation}
\tilde{h} = \left(1+\gamma(z_{ij})\right)\odot h + \beta(z_{ij}),
\label{eq:film}
\end{equation}
where $h$ denotes the intermediate feature tensor, $\gamma(\cdot)$ and $\beta(\cdot)$ are learned channel-wise scale and shift parameters, and $\odot$ denotes element-wise multiplication. The backbone is a four-level U-Net with skip connections \cite{ronneberger2015u}, using channel widths $[32,64,128,256]$ and a 512-channel bottleneck. FiLM-conditioned double-convolution blocks are used throughout to suppress noise while preserving fringe structure. In each decoder stage, features are upsampled, concatenated with the corresponding skip connection, and refined by a FiLM-conditioned double-convolution block. The network has two output heads: one predicts a restored complex interferogram in two-channel real/imaginary form, $\hat{y}_{ij}\in\mathbb{R}^{2\times H\times W}$, and the other predicts a per-pixel log-variance map for aleatoric uncertainty estimation, $\log \hat{\sigma}^{2}_{ij}\in\mathbb{R}^{1\times H\times W}$.

\subsection{Pseudo-Supervised Training Objective}

Training is formulated as geometry-conditioned pseudo-supervised regression, with Goldstein-filtered interferograms used as pseudo-targets. The model therefore learns a geometry-adaptive generalization of the classical Goldstein baseline without using reference DEMs or externally provided deformation labels. For each tile, FiLM-GPNet takes the raw interferogram and coherence channels as input and predicts a restored two-channel complex interferogram together with a per-pixel log-variance map. The total loss combines a wrapped-phase regression term, a heteroscedastic uncertainty loss, a fringe-preserving gradient loss, a temporal-consistency term and a triplet closure-consistency term:
\begin{equation}
\mathcal{L}
=
\lambda_{\mathrm{reg}} \mathcal{L}_{\mathrm{reg}}
+
\lambda_{\mathrm{unc}} \mathcal{L}_{\mathrm{unc}}
+
\lambda_{\mathrm{cls}} \mathcal{L}_{\mathrm{cls}}
+
\lambda_{\mathrm{tmp}} \mathcal{L}_{\mathrm{tmp}}
+
\lambda_{\mathrm{grd}} \mathcal{L}_{\mathrm{grd}}
\end{equation}
Here, $\mathcal{L}_{\mathrm{reg}}$ matches the prediction to the Goldstein pseudo-target, $\mathcal{L}_{\mathrm{unc}}$ models heteroscedastic uncertainty, $\mathcal{L}_{\mathrm{grd}}$ preserves fringe gradients, $\mathcal{L}_{\mathrm{cls}}$ enforces triplet closure, and $\mathcal{L}_{\mathrm{tmp}}$ promotes stack-level temporal consistency through an SBAS-style residual constraint. The scalar weights $\lambda_{\mathrm{reg}}$, $\lambda_{\mathrm{unc}}$, $\lambda_{\mathrm{cls}}$, $\lambda_{\mathrm{tmp}}$, and $\lambda_{\mathrm{grd}}$ balance the five terms. Overall, the objective defines a pseudo-supervised restoration model regularized by uncertainty and interferometric physics.

\begin{figure*}[t!]
  \centering
    \includegraphics[width=0.85\textwidth]{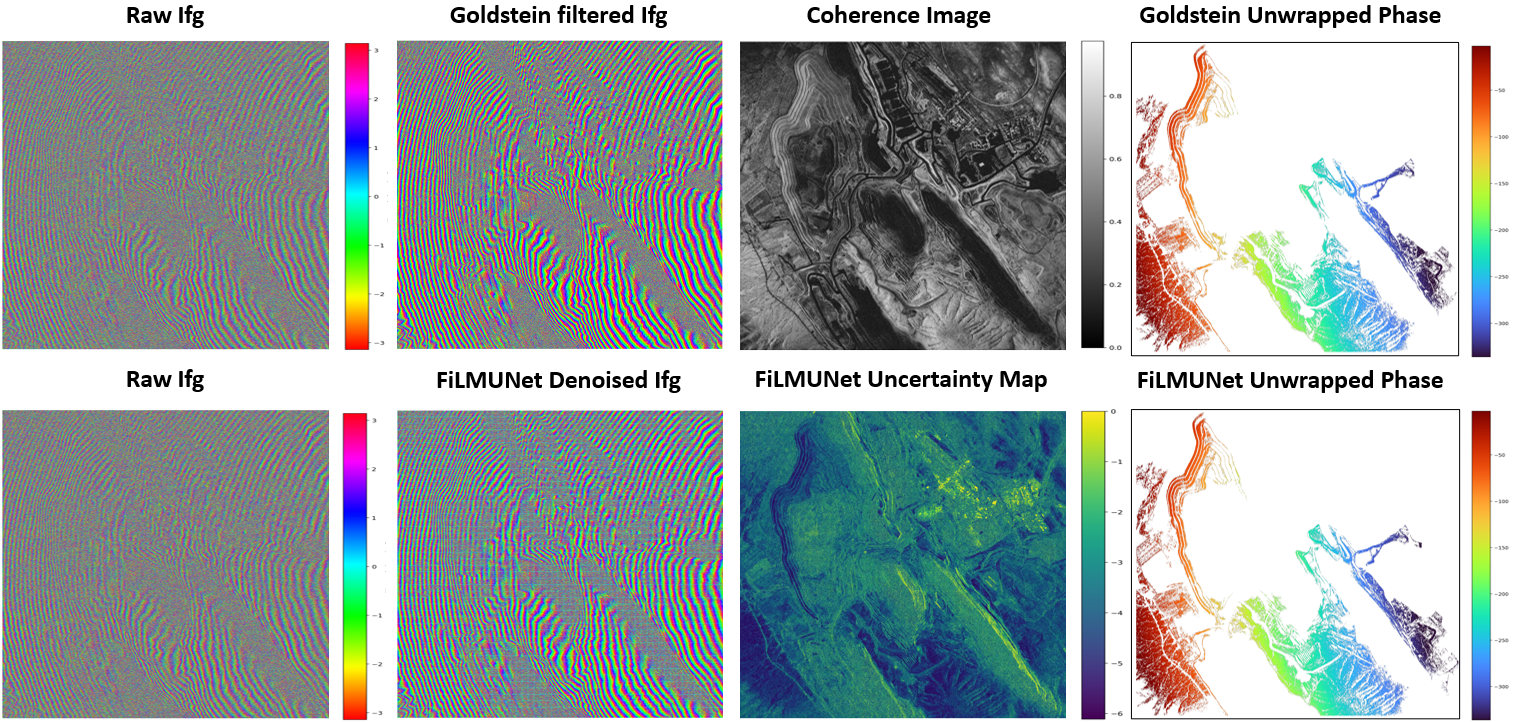}
    \caption{Visual Comparison of intermediate outputs. Top row: baseline InSAR outputs, Bottom row: FiLM-GPNet derived outputs.}
    \label{main-results}
\end{figure*}

\begin{figure*}[t!]
  \centering
    \includegraphics[width=0.85\textwidth]{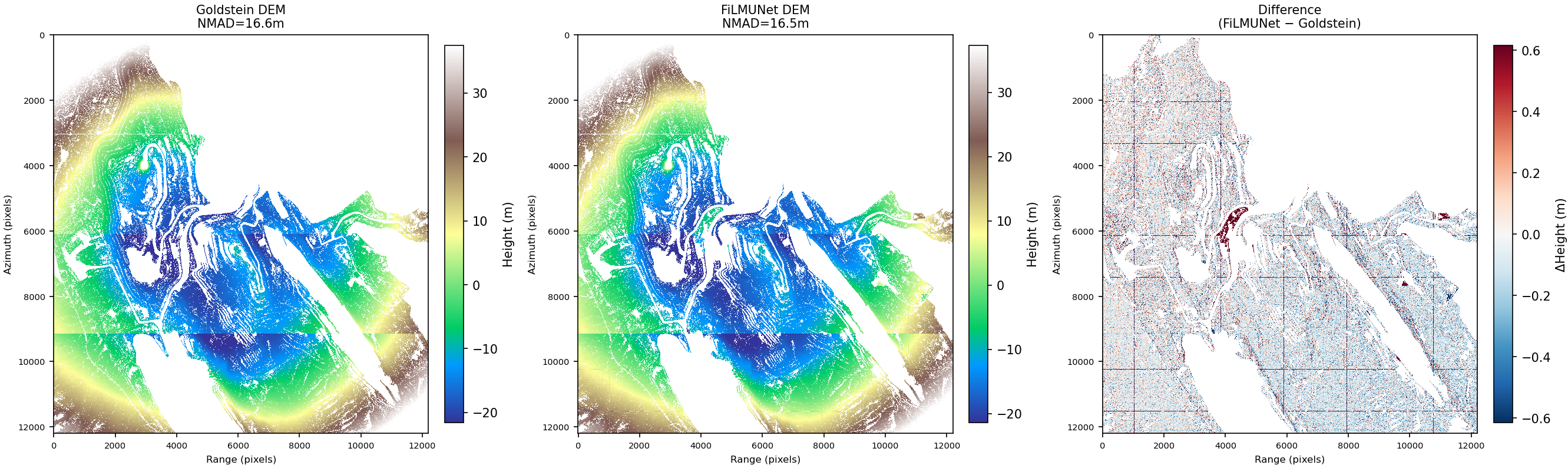}
    \caption{Comparison of elevation-domain outputs in native SAR range–azimuth grid, before geocoding: baseline vs FiLM-GPNet.}
    \label{DEM-Comp}
\end{figure*}

\subsection{Uncertainty-Aware Downstream Processing}

After tile-wise inference and overlap-add reconstruction, the FiLM-GPNet-denoised interferogram is processed by a standard InSAR back-end: the phase is unwrapped with SNAPHU, converted to elevation using the standard phase-to-height relation \cite{Hanssen2001}, and terrain-geocoded for comparison with external DEMs. These steps are used to evaluate unwrapping stability and elevation recovery. Algorithm~1 in the appendix summarizes the complete workflow, including geometry-aware preprocessing, the classical InSAR baseline, FiLM-GPNet-based phase denoising, and downstream post-processing.

\section{Experiments and Results}

\subsection{Data and Experimental Setup}

The IEEE GRSS 2026 Data Fusion Contest dataset~\cite{ieeegrss2026dfc} comprises $791$ Capella Space X-band Spotlight SLCs across $39$ AOIs. We use three stacks with distinct temporal and geometric characteristics: \textbf{AOI\_000} (Hawaii), \textbf{AOI\_008} (Los Angeles), and \textbf{AOI\_024} (Western Australia). AOI\_000 is the densest stack, with $221$ acquisitions spanning June 2024 to November 2025, mixed ascending/descending passes, and incidence angles of $35.8^\circ$--$56.3^\circ$. AOI\_024 is more homogeneous, with $100$ descending right-looking acquisitions and incidence angles of $34.5^\circ$--$39.4^\circ$, while AOI\_008 contains $119$ ascending right-looking acquisitions over a longer temporal span, with incidence angles of $35.3^\circ$--$41.4^\circ$. Together, these stacks provide a dense geometry-diverse primary evaluation site and two additional stacks for cross-stack and zero-shot assessment.

After strict pair-graph filtering, we retain geometrically compatible pairs and valid closure triplets within each AOI. Training uses an AOI-aware temporal split by acquisition date, and all metrics are reported on processed test stacks using $256\times256$ tiles with stride $128$. The model is trained with AdamW (${\rm lr}=10^{-4}$, weight decay $10^{-5}$), cosine annealing with a 2-epoch warm-up, batch size $16$, and $50$ epochs, on a single NVIDIA~A100 (24\,GB VRAM).

\subsection{Evaluation Metrics}

We evaluate performance using four complementary metrics that capture pairwise consistency, downstream usability, elevation quality, and stack-level temporal stability. These include \emph{Triplet Closure Error}, \emph{Unwrap Success Rate}, \emph{DEM NMAD}, and \emph{Temporal Residual}. Their formal definitions and equations are provided in the appendix.

\subsection{Results}

Table~\ref{tab-1} summarizes the four-metric evaluation protocol. FiLM-GPNet was trained on processed pairs from the Hawaii and Western Australia stacks (AOI\_000 and AOI\_024), while the Los Angeles stack (AOI\_008) was reserved for zero-shot evaluation. For both AOI\_000 and AOI\_024, FiLM-GPNet surpasses the classical Goldstein baseline on all four metrics. In particular, the temporal residual is reduced by $68\%$ and $66\%$, respectively, while triplet closure error decreases by $10\%$ and $13\%$. The disparity between these gains is itself informative: geometry-aware denoising improves \emph{stack-level} temporal coherence more strongly than pairwise consistency alone. For AOI\_024, FiLM-GPNet also improves unwrap success rate by $7.7$ percentage points and DEM NMAD by $31\%$, indicating that the restored phase remains fully usable for downstream unwrapping and elevation recovery. In the zero-shot setting on AOI\_008, FiLM-GPNet achieves nearly identical statistics to the classical Goldstein baseline, supporting its ability to generalize across geographically and geometrically distinct stacks without retraining.

\begin{table}[htbp]
\centering
\caption{Quantitative performance comparison across AOIs}
\label{tab:single_col_fit}
\setlength{\tabcolsep}{2pt} 
\scriptsize 
\begin{tabular}{lcccccc}
\toprule
\textbf{Metric} & \multicolumn{3}{c|}{\textbf{Goldstein AOIs}} & \multicolumn{3}{c}{\textbf{FiLM-GPNet AOIs ($\Delta$)}} \\
\cmidrule(lr){2-4} \cmidrule(lr){5-7}
& \textbf{000} & \textbf{024} & \textbf{008} & \textbf{000} & \textbf{024} & \textbf{008} \\
\midrule
Closure $\downarrow$ & 1.018 & 0.536 & 0.769 & \textbf{0.915} (-10\%) & \textbf{0.468} (-13\%) & 0.771 (+0.3\%) \\
Unwrap $\uparrow$ & 0.256 & 0.531 & 0.256 & \textbf{0.258} (+.2 pp) & \textbf{0.608} (+7.7 pp) & 0.248 (-.8 pp) \\
NMAD $\downarrow$  & 40.13 & 18.32 & 40.13 & \textbf{39.44} (-2\%) & \textbf{12.64} (-31\%) & \textbf{39.40} (-2\%) \\
Temp. R. $\downarrow$ & 1.158 & 1.069 & 1.486 & \textbf{0.367} (-68\%) & \textbf{0.361} (-66\%) & \textbf{1.450} (-2\%) \\
\bottomrule
\label{tab-1}
\end{tabular}
\end{table}

\begin{table}[t]
\centering
\caption{Loss and architecture ablation on the Hawaii sub-stack.}
\label{tab:ablation}
\setlength{\tabcolsep}{4pt}
\renewcommand{\arraystretch}{1.0}
\footnotesize
\begin{tabular}{lccccc}
\toprule
\textbf{Variant} & $\mathcal{L}_{\rm cls}$ & $\mathcal{L}_{\rm tmp}$ & \textbf{FiLM} & \textbf{Closure$\downarrow$} & \textbf{Temp.$\downarrow$} \\
\midrule
Pseudo-Supervised (P-S) & \texttimes & \texttimes & \checkmark & 1.021 & 0.653 \\
P-S + $\mathcal{L}_{\rm cls}$ + $\mathcal{L}_{\rm tmp}$ & \checkmark & \checkmark & \checkmark & \textbf{0.915} & \textbf{0.367} \\
\bottomrule
\end{tabular}
\end{table}

Table~\ref{tab:ablation} reports the ablation study on the Hawaii stack. Removing the closure and temporal-consistency terms degrades both closure error and temporal residual, confirming that the physics-based losses, rather than the pseudo-supervised target alone, account for most of the observed gains. Figure~\ref{main-results} presents representative intermediate outputs and visual comparisons between FiLM-GPNet and the classical baseline. Figure~\ref{DEM-Comp} compares the elevation-domain outputs of the classical baseline and FiLM-GPNet on the native SAR range–azimuth grid, prior to geocoding. Additional qualitative results for the zero-shot experiment on AOI\_008 are provided in the appendix.

\section{Conclusion}


FiLM-GPNet shows that explicitly conditioning phase restoration on acquisition geometry, combined with physics-informed pseudo-supervision, enables more consistent and robust processing of dense temporal InSAR stacks than fixed classical filtering approaches. Across heterogeneous Capella Spotlight stacks, the proposed method improves temporal and stack-level phase consistency, with the most significant gains observed in temporal residual reduction on the Hawaii and Western Australia datasets, while maintaining downstream performance in phase unwrapping and DEM quality. Importantly, the model generalizes in a zero-shot setting to a geographically and geometrically distinct stack without retraining, indicating that geometry-conditioned restoration provides a scalable and transferable alternative to fixed filtering strategies for large commercial SAR archives.

\section*{Acknowledgment}
This research was supported by the Ministry of Education and Culture’s Doctoral Education Pilot under Decision No. VN/3137/2024-OKM-6, as part of the Digital Waters (DIWA) Doctoral Education Pilot, associated with the DIWA Flagship funded by the Research Council of Finland’s Flagship Programme (Decisions Nos. 359247 (University of Turku) and 359228 (University of Oulu)).

\small
\bibliographystyle{IEEEtranN}
\bibliography{references}

\end{document}


\maketitle

\renewcommand{\thefigure}{A\arabic{figure}}
\setcounter{figure}{0}
\section{End-to-End Pipeline Overview}
\begin{figure*}[!t]
  \centering
  \includegraphics[width=\textwidth]{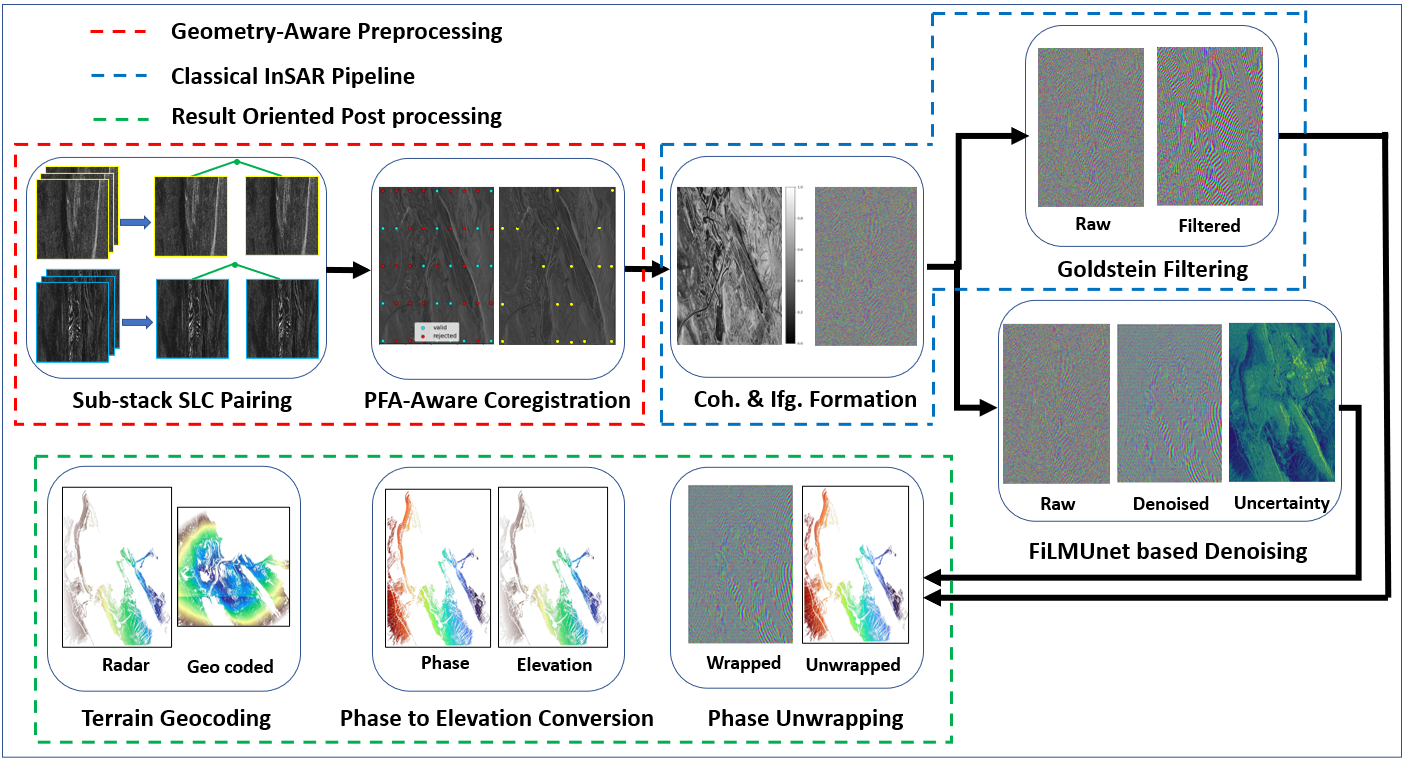}
  \caption{Complete InSAR Processing Pipeline: (a) geometry-aware preprocessing, (b) two main branches --- (b1) classical Goldstein baseline and (b2) FiLM-GPNet learned denoising --- and (c) standard post-processing steps including unwrapping, DEM inversion, and SBAS stack analysis.}
  \label{pipeline}
\end{figure*}

Figure~\ref{pipeline} shows the complete FiLM-GPNet processing pipeline. It consists of three stages: (a) geometry-aware preprocessing (pair-graph construction, coregistration, interferogram formation, Goldstein filtering), (b) two parallel branches --- a classical Goldstein baseline (b1) and the learned FiLM-GPNet denoising branch (b2) --- and (c) standard InSAR post-processing (SNAPHU unwrapping, phase-to-height conversion, geocoding, SBAS inversion).

Algorithm~\ref{alg:pipeline} provides the step-by-step pseudocode for the full workflow.

\begin{algorithm}[!t]
\footnotesize
\caption{Geometry-Aware Pseudo-Supervised FiLM-GPNet Pipeline}
\label{alg:pipeline}
\begin{algorithmic}[1]
\Require Manifest of $N$ Capella Spotlight SLC acquisitions
\Ensure Denoised interferograms $\{\hat{I}_{ij}\}$, uncertainty maps $\{\log \hat{\sigma}^{2}_{ij}\}$, and downstream unwrapped/geocoded products
\State Construct a strict interferometric pair graph using geometry-aware compatibility constraints; rank accepted edges by $Q_{ij}$
\State Enumerate valid closure triplets from the accepted graph
\ForAll{accepted pairs $(i,j)$}
    \State Coregister the slave SLC to the master using metadata seeding, thumbnail phase correlation, and local tie-point refinement
    \State Select the best coregistration candidate by coherence statistics
    \State Form raw interferogram $I_{ij}$ and coherence map $\gamma_{ij}$
    \State Reject pair if mean coherence $\bar{\gamma}_{ij}<0.3$
    \State Generate Goldstein-filtered pseudo-target $I^{\mathrm{G}}_{ij}$
    \State Assemble normalized geometry vector
    \[
    \mathbf{z}_{ij}=[\Delta t_{ij},\theta_{\mathrm{inc},ij},\theta_{\mathrm{graze},ij},B_{\perp,ij},m^{\mathrm{mode}}_{ij},m^{\mathrm{look}}_{ij},\mathrm{SNR}_{ij}]
    \]
    \State Extract overlapping input tiles $(\Re(I_{ij}),\Im(I_{ij}),\gamma_{ij})$ and pseudo-target tiles $(\Re(I^{\mathrm{G}}_{ij}),\Im(I^{\mathrm{G}}_{ij}))$
\EndFor
\State \textbf{Train} FiLM-GPNet $f_{\theta}$ by minimizing
\[
\mathcal{L}
=
\lambda_{\mathrm{reg}}\mathcal{L}_{\mathrm{reg}}
+
\lambda_{\mathrm{unc}}\mathcal{L}_{\mathrm{unc}}
+
\lambda_{\mathrm{cls}}\mathcal{L}_{\mathrm{cls}}
+
\lambda_{\mathrm{grd}}\mathcal{L}_{\mathrm{grd}},
\]
\Statex \qquad where $\mathcal{L}_{\mathrm{reg}}$ uses Goldstein pseudo-targets and $\mathcal{L}_{\mathrm{cls}}$ is applied when valid triplet batches are available
\ForAll{evaluation pairs $(i,j)$}
    \State Predict tile-wise outputs
    \[
    \hat{I}_{ij},\,\log \hat{\sigma}^{2}_{ij} \gets f_{\theta}(I_{ij},\gamma_{ij},\mathbf{z}_{ij})
    \]
    \State Reconstruct full-resolution $\hat{I}_{ij}$ and $\log \hat{\sigma}^{2}_{ij}$ by overlap-add fusion
    \State Extract wrapped phase from $\hat{I}_{ij}$, unwrap with SNAPHU, convert phase to elevation, and terrain-geocode the result
\EndFor
\State Retain $\log \hat{\sigma}^{2}_{ij}$ as an auxiliary confidence layer for screening and SBAS-compatible downstream processing
\end{algorithmic}
\end{algorithm}

\section{FiLM-GPNet Architecture Details}

FiLM-GPNet is a four-scale encoder-decoder network with skip connections, adapted from the U-Net family~\cite{ronneberger2015u}. The input is a three-channel tensor $x_{ij} \in \mathbb{R}^{3 \times H \times W}$ formed by concatenating the real part, imaginary part, and coherence map of the Goldstein-filtered interferogram. The architecture uses channel widths $[32, 64, 128, 256]$ at the four encoder scales and a 512-channel bottleneck.

\paragraph{Geometry conditioning via FiLM.}
Each pair is associated with a seven-dimensional geometry vector
\begin{equation}
\mathbf{z}_{ij} = [\Delta t_{ij},\, \theta_{\mathrm{inc},ij},\, \theta_{\mathrm{graze},ij},\, B_{\perp,ij},\, m_{ij}^{\mathrm{mode}},\, m_{ij}^{\mathrm{look}},\, \mathrm{SNR}_{ij}],
\end{equation}
which is z-scored at the dataset level. A two-layer MLP maps this 7-D input to a 64-D conditioning code $\mathbf{c} \in \mathbb{R}^{64}$. At each encoder and decoder block, FiLM applies channel-wise affine modulation to the intermediate feature map $h$:
\begin{equation}
\tilde{h} = \left(1 + \gamma(\mathbf{c})\right) \odot h + \beta(\mathbf{c}),
\end{equation}
where $\gamma(\cdot)$ and $\beta(\cdot)$ are linear projections from $\mathbb{R}^{64}$ to the number of channels in $h$. The residual form $(1 + \gamma)$ initializes close to the identity, ensuring stable early training. FiLM is applied after each batch normalization layer within the double-convolution blocks.

\paragraph{Output heads.}
The network has two output heads: (1) a restored complex interferogram $\hat{y}_{ij} \in \mathbb{R}^{2 \times H \times W}$ (real and imaginary channels) and (2) a per-pixel log-variance map $\log \hat{\sigma}^{2}_{ij} \in \mathbb{R}^{1 \times H \times W}$ for aleatoric uncertainty estimation. The total parameter count is approximately 7.96~M.

\section{Loss Function Details}

The total training loss is a weighted sum of four terms:
\begin{equation}
\mathcal{L} = \lambda_{\mathrm{reg}}\mathcal{L}_{\mathrm{reg}} + \lambda_{\mathrm{unc}}\mathcal{L}_{\mathrm{unc}} + \lambda_{\mathrm{cls}}\mathcal{L}_{\mathrm{cls}} + \lambda_{\mathrm{grd}}\mathcal{L}_{\mathrm{grd}}.
\end{equation}

\paragraph{Pseudo-supervised regression loss $\mathcal{L}_{\mathrm{reg}}$.}
The primary supervision signal comes from Goldstein-filtered interferograms used as pseudo-targets. For each tile, the L1 loss is computed on the wrapped phase of the prediction versus the Goldstein output:
\begin{equation}
\mathcal{L}_{\mathrm{reg}} = \frac{1}{HW} \sum_{p} \left| \mathcal{W}\!\left(\angle\hat{I}(p) - \angle I^{\mathrm{G}}(p)\right) \right|,
\end{equation}
where $\mathcal{W}(\cdot)$ wraps differences to $[-\pi, \pi)$. This encourages the model to generalize the classical Goldstein response in a geometry-adaptive way, going beyond what a fixed $\alpha$ parameter can achieve. Weight: $\lambda_{\mathrm{reg}} = 1.0$.

\paragraph{Heteroscedastic uncertainty loss $\mathcal{L}_{\mathrm{unc}}$.}
The model is trained to calibrate its own uncertainty via a negative log-likelihood (NLL) objective under a Laplacian noise model:
\begin{equation}
\mathcal{L}_{\mathrm{unc}} = \frac{1}{HW} \sum_{p} \left( \frac{|\mathcal{W}(\angle\hat{I}(p) - \angle I^{\mathrm{G}}(p))|}{\exp(\frac{1}{2}\log\hat{\sigma}^2(p))} + \frac{1}{2}\log\hat{\sigma}^2(p) \right).
\end{equation}
Pixels with high residual incur less penalty if $\hat{\sigma}^2$ is large, while the regularization term prevents trivial solutions. This yields spatially calibrated uncertainty: high $\hat{\sigma}^2$ concentrates on incoherent and geometrically unfavorable pixels. Weight: $\lambda_{\mathrm{unc}} = 0.5$.

\paragraph{Triplet closure consistency loss $\mathcal{L}_{\mathrm{cls}}$.}
When a batch contains a valid triplet $(i, j, k)$ --- i.e., all three interferograms $(i,j)$, $(j,k)$, $(i,k)$ are present --- the closure residual is computed on the predicted phases and penalized:
\begin{equation}
\mathcal{L}_{\mathrm{cls}} = \frac{1}{|\mathcal{T}|} \sum_{(i,j,k)\in\mathcal{T}} \frac{1}{HW} \sum_{p} \left| \mathcal{W}\!\left(\angle\hat{I}_{ij}(p) + \angle\hat{I}_{jk}(p) - \angle\hat{I}_{ik}(p)\right) \right|.
\end{equation}
This enforces the fundamental InSAR closure constraint directly on the model predictions, encouraging stack-level self-consistency beyond what pairwise supervision alone provides. Weight: $\lambda_{\mathrm{cls}} = 0.3$.

\paragraph{Gradient preservation loss $\mathcal{L}_{\mathrm{grd}}$.}
To prevent over-smoothing of phase fringes, a gradient-domain L1 loss is applied between the predicted and pseudo-target phase:
\begin{equation}
\mathcal{L}_{\mathrm{grd}} = \frac{1}{HW} \sum_{p} \left( |\nabla_x \hat{\phi}(p) - \nabla_x \phi^{\mathrm{G}}(p)| + |\nabla_y \hat{\phi}(p) - \nabla_y \phi^{\mathrm{G}}(p)| \right),
\end{equation}
where $\hat{\phi} = \angle\hat{I}$ and $\phi^{\mathrm{G}} = \angle I^{\mathrm{G}}$. This term anchors the spatial frequency content of the prediction to that of the Goldstein target, preserving sharp fringe boundaries. Weight: $\lambda_{\mathrm{grd}} = 0.1$.

\section{Training and Implementation Details}

\paragraph{Data partitioning.}
For each AOI, pairs are sorted by acquisition date and split 70/15/15\% (train/val/test) at the collect level to prevent temporal leakage. All evaluation metrics are reported on the test split only.

\paragraph{Tile extraction.}
Interferograms are tiled at $256 \times 256$ pixels with a stride of 128 pixels (50\% overlap). Tiles with fewer than 10\% valid (non-NaN, non-zero-coherence) pixels are discarded. For AOI\_000 (Hawaii) this yields approximately 150,000 training tiles.

\paragraph{Optimizer and schedule.}
AdamW optimizer with learning rate $10^{-4}$ and weight decay $10^{-5}$. Cosine annealing schedule with a 2-epoch linear warm-up. Batch size 8. Training runs for 50 epochs on a single NVIDIA A100 (40 GB). Random seed 42 throughout.

\paragraph{Inference.}
At inference time, tiles are extracted with the same $256 \times 256$ / stride-128 configuration and the full-resolution output is reconstructed by overlap-add averaging. Inference runs at approximately 0.2\,s per pair on GPU and 3\,s per pair on CPU, scaling to the full 791-SLC contest dataset in under 15\,min on a single GPU.

\paragraph{Zero-shot transfer.}
For AOI\_008 (Los Angeles), the model checkpoint trained on AOI\_024 (Western Australia) is applied directly without any fine-tuning or adaptation. The only operation performed is geometry vector normalization using the AOI\_024 training statistics (mean and standard deviation), which is a fixed pre-processing step requiring no labeled data from the target site.

\paragraph{SBAS inversion.}
The SBAS design matrix $\mathbf{A} \in \{-1, 0, +1\}^{P \times T}$ encodes the pair-to-epoch mapping. Weighted least squares (WLS) inversion uses diagonal weights $W_{ij} = 1 / \bar{\sigma}^2_{ij}$, where $\bar{\sigma}^2_{ij}$ is the spatially averaged predicted variance for pair $(i,j)$. The temporal residual metric is computed using the unweighted residual to allow fair comparison with Goldstein.

\section{Per-AOI Dataset Statistics}

\begin{table}[H]
\centering
\caption{Per-AOI dataset statistics used in this work.}
\label{tab:aoi_stats}
\setlength{\tabcolsep}{3pt}
\footnotesize
\resizebox{\columnwidth}{!}{%
\begin{tabular}{lccc}
\toprule
\textbf{Property} & \textbf{AOI\_000 (Hawaii)} & \textbf{AOI\_024 (W.\ Australia)} & \textbf{AOI\_008 (Los Angeles)} \\
\midrule
Total acquisitions     & 221   & 100   & 119  \\
Orbit direction        & Asc + Desc & Desc only & Asc only \\
Look direction         & Right & Right & Right \\
Incidence range (°)    & 35.8--56.3 & 34.5--39.4 & 35.3--41.4 \\
Processed pairs        & 224   & 4     & 4 (zero-shot) \\
Valid closure triplets & 62    & ---   & --- \\
Training role          & Primary & Fine-tune & Zero-shot eval \\
\bottomrule
\end{tabular}%
}
\end{table}

\section{Evaluation Metric Definitions}

All four metrics are computed on the test split of each AOI. Goldstein and FiLM-GPNet predictions are processed through identical post-processing pipelines (unwrapping, SBAS inversion) to ensure a fair comparison.

\subsection{Triplet Closure Error (M1)}

The closure error measures interferometric phase consistency over valid triplets $(i,j,k)$ formed from three acquisitions $i < j < k$:
\begin{equation}
\varepsilon_{ijk}(p) = \mathcal{W}\!\left(\phi_{ij}(p) + \phi_{jk}(p) - \phi_{ik}(p)\right),
\end{equation}
where $\phi_{ij}$ is the wrapped phase of pair $(i,j)$ and $\mathcal{W}(\cdot)$ wraps to $[-\pi,\pi)$. For a perfectly consistent set of interferograms, this residual is identically zero everywhere. The reported metric is the mean absolute wrapped closure residual, averaged over all valid pixels and all valid triplets:
\begin{equation}
\mathrm{M1} = \frac{1}{|\mathcal{T}|} \sum_{(i,j,k)\in\mathcal{T}} \frac{1}{|\Omega_{ijk}|} \sum_{p \in \Omega_{ijk}} |\varepsilon_{ijk}(p)|.
\end{equation}
Lower values indicate more self-consistent interferogram stacks. Systematic decorrelation or geometry-dependent phase errors inflate closure error even when individual interferograms appear clean.

\subsection{Unwrap Success Rate (M2)}

The unwrap success rate quantifies the fraction of valid interferometric pixels that SNAPHU successfully assigns a reliable unwrapped phase value:
\begin{equation}
\mathrm{M2} = \frac{1}{|\mathcal{P}|} \sum_{(i,j)\in\mathcal{P}} \frac{|\Omega^{\mathrm{unwrap}}_{ij}|}{|\Omega^{\mathrm{valid}}_{ij}|},
\end{equation}
where $\Omega^{\mathrm{valid}}_{ij}$ is the set of pixels passing the coherence threshold before unwrapping, and $\Omega^{\mathrm{unwrap}}_{ij}$ is the subset assigned reliable unwrapped phase by SNAPHU's connected-component algorithm. Higher values indicate that denoised interferograms are more amenable to phase unwrapping. For X-band data over densely vegetated or geologically rough terrain (e.g., Hawaii), the absolute success rate is inherently low due to scene-level decorrelation independent of the filtering method.

\subsection{DEM NMAD (M3)}

DEM quality is assessed using the normalized median absolute deviation (NMAD) between the InSAR-derived elevation $\hat{h}$ and a Copernicus GLO-30 reference DEM $h^{\mathrm{ref}}$. For each valid comparison pixel $p \in \Omega_h$, the height error is:
\begin{equation}
e_h(p) = \hat{h}(p) - h^{\mathrm{ref}}(p).
\end{equation}
NMAD is then computed as:
\begin{equation}
\mathrm{M3} = 1.4826 \cdot \mathrm{median}_{p \in \Omega_h}\!\left(\left|e_h(p) - \mathrm{median}(e_h)\right|\right).
\end{equation}
The factor 1.4826 normalizes NMAD to be a consistent estimator of the standard deviation under a Gaussian distribution. NMAD is preferred over RMSE because it is robust to large outliers arising from phase unwrapping errors at incoherent pixels. The valid comparison domain $\Omega_h$ excludes pixels flagged as invalid by SNAPHU and those outside the reference DEM coverage.

\subsection{Temporal Residual (M4)}

The temporal residual measures stack-level consistency between the processed interferogram sequence and the best-fitting SBAS temporal model. Given the SBAS design matrix $\mathbf{A} \in \{-1,0,+1\}^{P \times T}$ and the stack of unwrapped phases $\hat{\boldsymbol{\phi}} \in \mathbb{R}^{P}$ at a given pixel, the SBAS solution $\mathbf{x}^* \in \mathbb{R}^T$ minimizes $\|\mathbf{A}\mathbf{x} - \hat{\boldsymbol{\phi}}\|_2^2$. The per-pixel wrapped residual is:
\begin{equation}
r_{\Phi}(p) = \mathcal{W}\!\left(\hat{\boldsymbol{\phi}}(p) - \mathbf{A}\mathbf{x}^{*}(p)\right).
\end{equation}
The reported metric is the mean absolute residual over all valid stack pixels:
\begin{equation}
\mathrm{M4} = \frac{1}{|\Omega_{\Phi}|} \left\| r_{\Phi} \right\|_{1}.
\end{equation}

\begin{figure*}[!t]
  \centering
  \includegraphics[width=\textwidth]{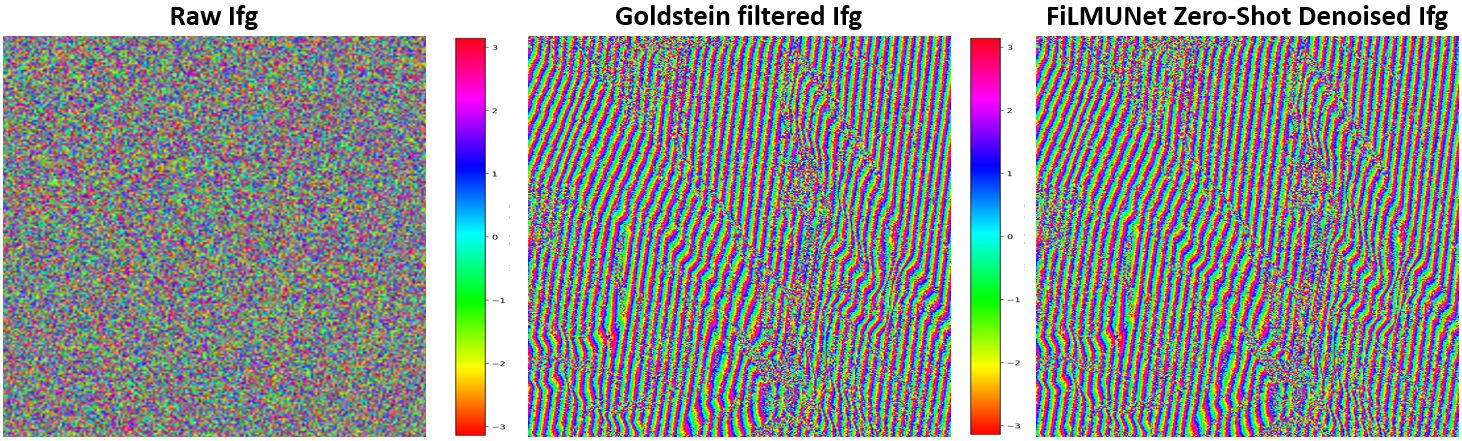}
  \caption{Visualization results of Zero-Shot transfer (Ifg denoising) on AOI008 by inference only, on model trained on the AOI024 stack.}
  \label{zero-shot}
\end{figure*}

The temporal residual aggregates phase errors across the entire interferogram stack, making it highly sensitive to geometry-dependent inconsistencies that cancel out or are invisible in pairwise metrics. It is computed using \emph{unweighted} SBAS inversion for fair comparison between Goldstein and FiLM-GPNet, even though FiLM-GPNet uncertainty weights are used during the actual SBAS inversion for downstream products.

\section{Results}
Figure \ref{zero-shot} shows the visualization results of Zero-Shot transfer (Ifg denoising) on AOI008 by performing inference only, on the model trained on the AOI024 stack. The result of Goldstein filtering is also shown for comparison.

\begin{figure}[!t]
\centering
\includegraphics[width=\columnwidth]{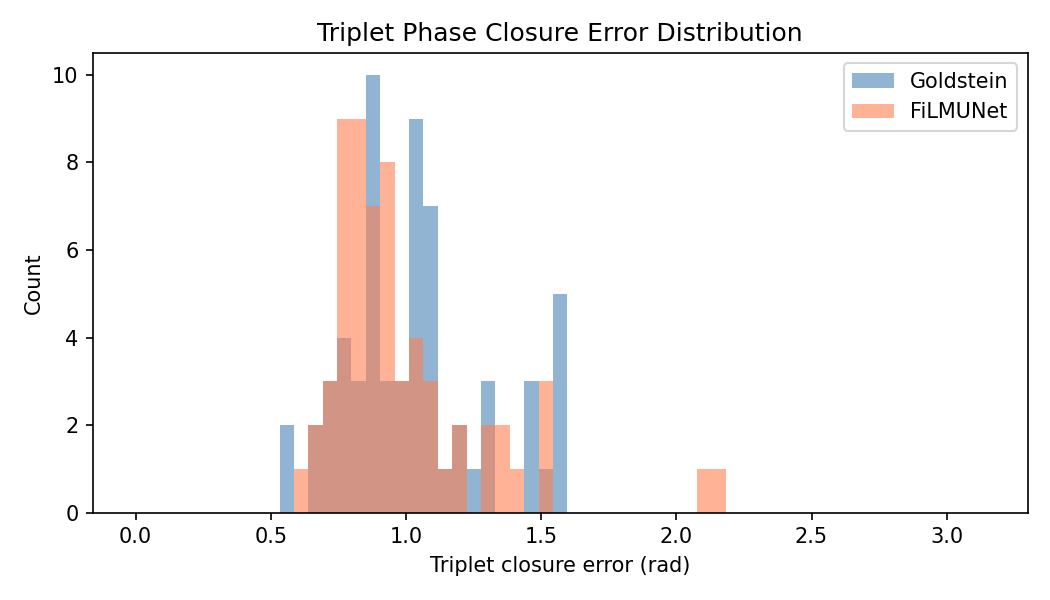}
\caption{Triplet closure error distribution across 62 Hawaii triplets.}
\label{fig:closure_hist}
\end{figure}

Figure \ref{fig:closure_hist} shows the triplet closure error distribution across 62 Hawaii triplets. FiLM-GPNet (coral) concentrates closer to zero than Goldstein (blue), indicating improved inter-pair phase consistency. Median: Goldstein 1.018\,rad, FiLM-GPNet 0.915\,rad ($-10.1\%$).

\begin{figure}[!t]
\centering
\includegraphics[width=5.6cm]{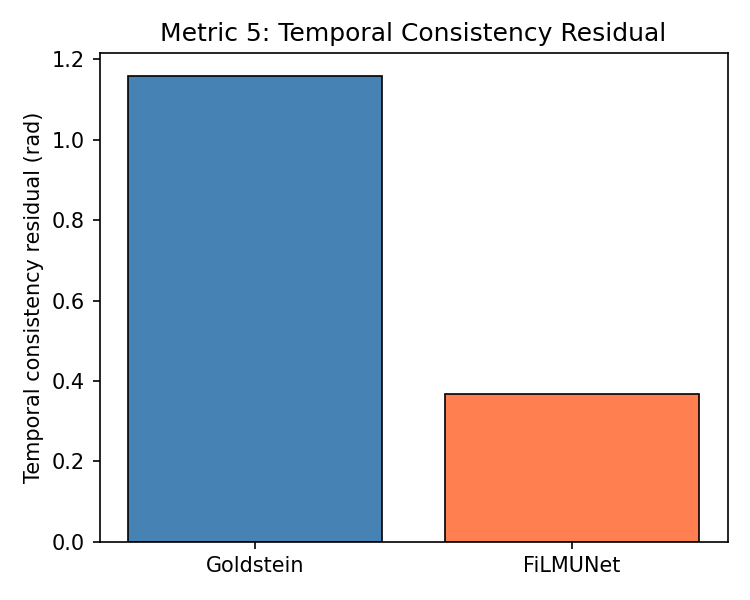}
\caption{SBAS temporal residual per pair.}
\label{fig:temporal_residual}
\end{figure}

Figure \ref{fig:temporal_residual} shows SBAS temporal residual per pair. FiLM-GPNet (coral) systematically reduces the unweighted residual $\|\hat{\boldsymbol{\varphi}} - \mathbf{A}\mathbf{x}^*\|$ across the stack, reflecting improved temporal self-consistency of the denoised interferogram sequence.

\begin{figure}[!t]
\centering
\includegraphics[width=7cm]{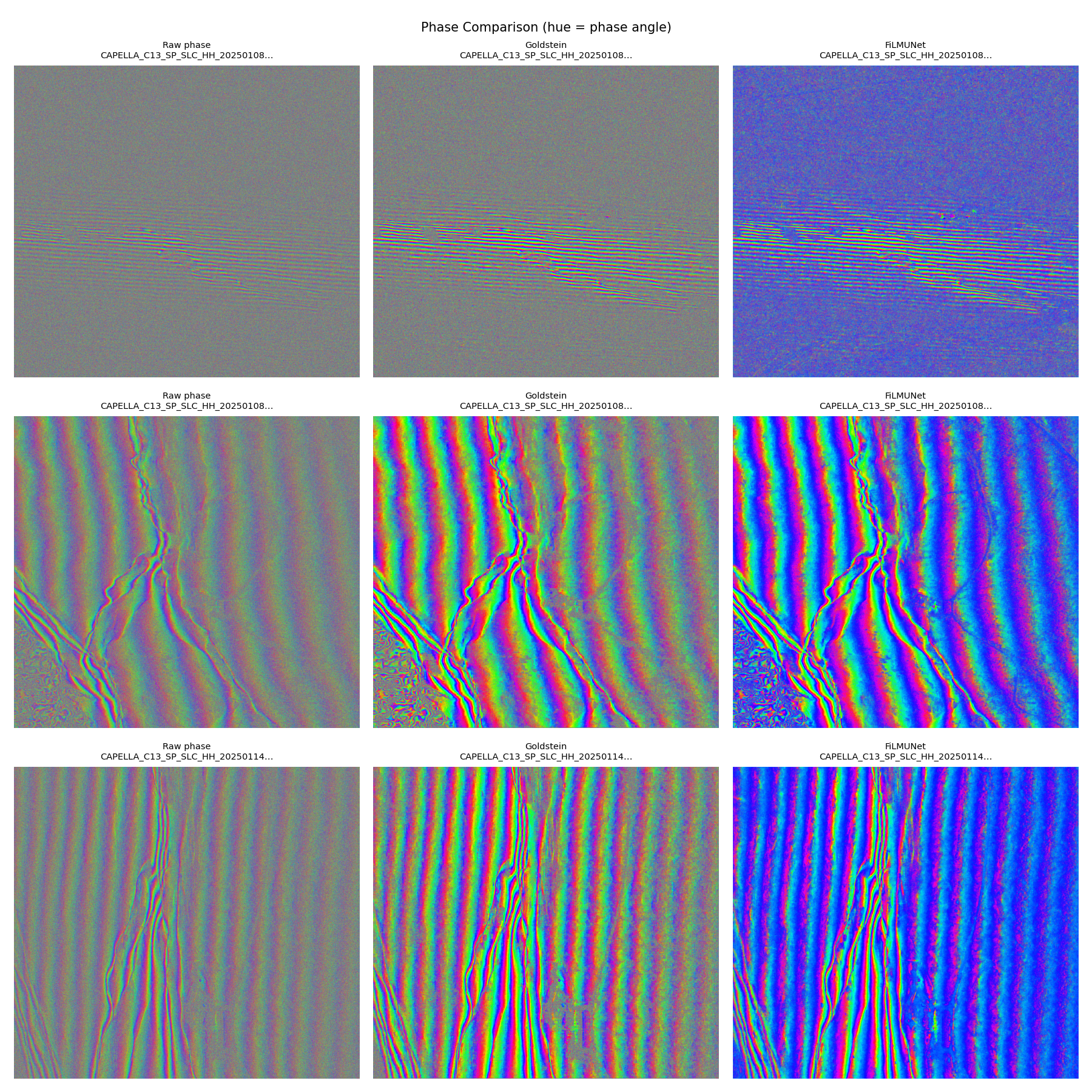}
\caption{Representative Hawaii pair (active lava field, Big Island).}
\label{fig:phase_comparison}
\end{figure}

Figure \ref{fig:phase_comparison} shows the results on the representative Hawaii stack (active lava field, Big Island). Left to right: raw complex interferogram, Goldstein output, FiLM-GPNet output. FiLM-GPNet reduces speckle while preserving fringe sharpness; the predicted uncertainty map (far right, red = high) correctly highlights incoherent regions.

\begin{figure*}[t]
\centering
\includegraphics[width=\textwidth]{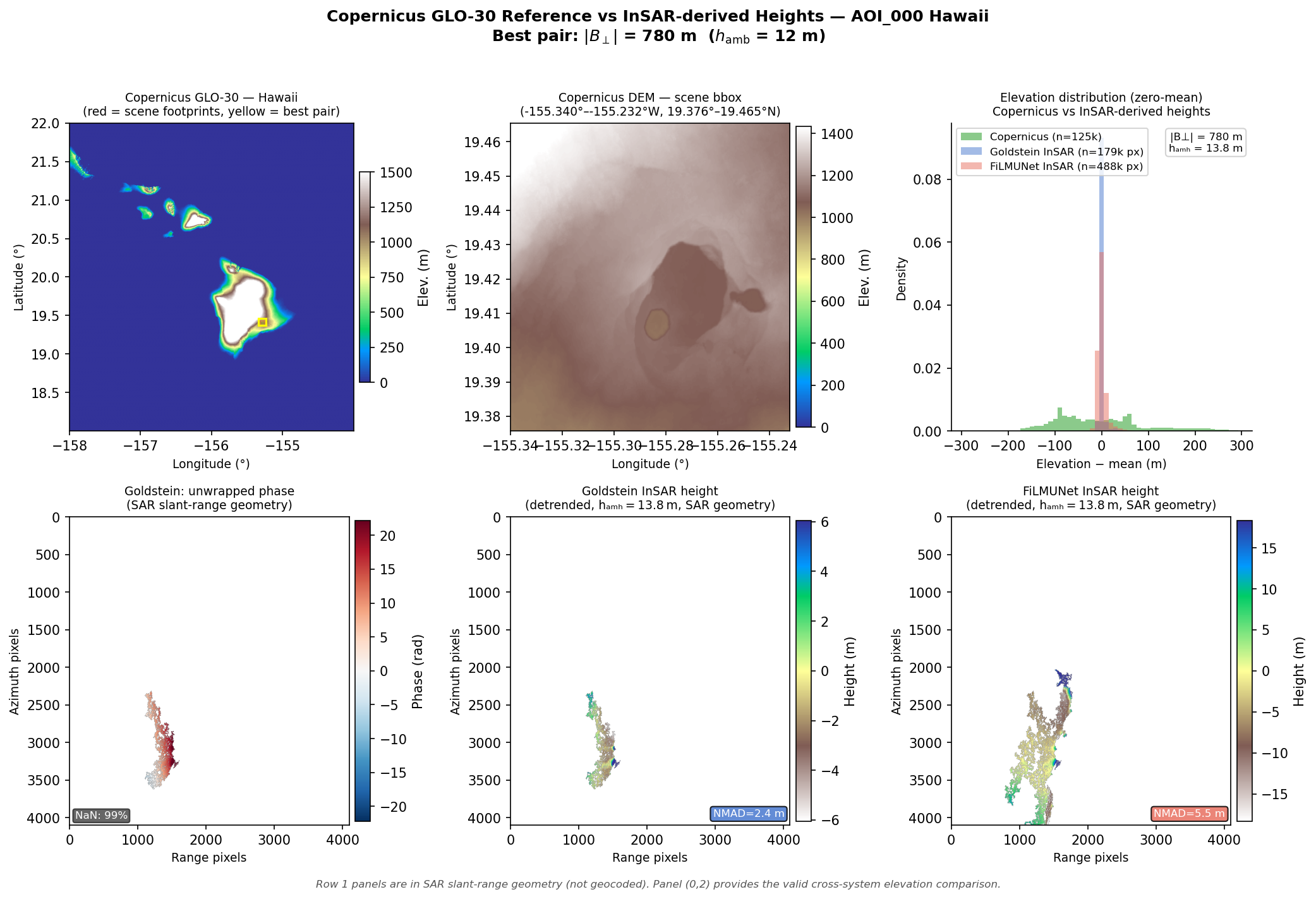}
\caption{Copernicus GLO-30 and InSAR-derived heights for the highest-sensitivity pair ($|B_\perp|=780$\,m, $h_{\rm amb}=12$\,m). Top: Hawaii overview, DEM subset, and elevation histogram. Bottom: Goldstein unwrapped phase, Goldstein height, and FiLM-GPNet height in SAR geometry.}
\label{fig:dem_comparison}
\end{figure*}

Figure \ref{fig:dem_comparison} compares the reference DEM with elevations recovered from the most height-sensitive interferometric pair.  
The top row provides geographic context and the cross-system elevation distribution.  
The bottom row shows the phase and derived height products in radar geometry for Goldstein and FiLM-GPNet.  
Visually, FiLM-GPNet preserves the main height structure while producing a cleaner elevation field.

\small
\bibliographystyle{IEEEtranN}
\bibliography{references}